\documentclass[11pt,a4paper]{article}
\usepackage[hyperref]{acl2020}
\usepackage{times}
\usepackage{latexsym}
\usepackage{graphicx}
\usepackage[capitalize,noabbrev]{cleveref}

\usepackage{microtype}

\aclfinalcopy % Uncomment this line for the final submission
\usepackage[normalem]{ulem} % sout
\title{Unsupervised Multilingual Sentence Embeddings \\
for Parallel Corpus Mining}

\author{Ivana Kvapil\'{i}kov\'{a}\textsuperscript{$\ddagger$}
        \quad Mikel Artetxe\textsuperscript{$\mathsection$} 
        \quad Gorka Labaka\textsuperscript{$\mathsection$} \\
       \quad \textbf{Eneko Agirre}\textsuperscript{$\mathsection$}   
       \quad \textbf{Ond\v{r}ej Bojar}\textsuperscript{$\ddagger$}
		\\ \\
\textsuperscript{$\ddagger$}Institute of Formal and Applied Linguistics, Charles University (MFF UK) \\
{\tt \{kvapilikova,bojar\}@ufal.mff.cuni.cz}\\
\textsuperscript{$\mathsection$}Ixa NLP group, University of the Basque Country (UPV/EHU)  \\
{\tt \{mikel.artetxe,gorka.labaka,e.agirre\}@ehu.eus} }

\date{}

\begin{document}
\maketitle
\begin{abstract}
%Recent work has shown that it is possible to create multilingual knowledge in models (e.g. XLM, M-BERT) trained on purely monolingual data. This paper proposes a method to enhance this knowledge 

%Multilingual sentence embeddings performant trained on parallel corpora yield impressive results e.g. on the task of parallel corpus mining. They 

Existing models of multilingual sentence embeddings require large parallel data resources which are not available for low-resource languages. We propose a novel unsupervised method to derive multilingual sentence embeddings relying only on monolingual data. We first produce a synthetic parallel corpus using unsupervised machine translation, and use it to fine-tune a pretrained cross-lingual masked language model (XLM) to derive the multilingual sentence representations. The quality of the representations is evaluated on two parallel corpus mining tasks with improvements of up to 22 F1 points over vanilla XLM. In addition, we observe that a single synthetic bilingual corpus is able to improve results for other language pairs.

%We fine-tune a pretrained cross-lingual masked language model (XLM) using a small synthetic parallel corpus obtained via unsupervised machine translation and derive sentence representations from contextualized word embeddings. We evaluate the embeddings on the task of parallel corpus mining from comparable corpora (BUCC) and report an improvement of 15.0 - 21.6 FI points over the vanilla XLM. % Our method is even comparable to several supervised baselines.

\end{abstract}

\section{Introduction}

Parallel corpora constitute an essential training data resource for machine translation as well as other cross-lingual NLP tasks. However, large parallel corpora are only available for a handful of language pairs while the rest relies on semi-supervised or unsupervised methods for training. Since monolingual data are generally more abundant, parallel sentence mining from non-parallel corpora provides another opportunity for low-resource language pairs. % is to obtain parallel sentences via parallel data mining.

An effective approach to parallel data mining is based on multilingual sentence embeddings \citep{schwenk-2018-filtering,Artetxe2019laser}. However, existing methods to generate cross-lingual representations are either heavily supervised or only apply to static word embeddings. An alternative approach to unsupervised multilingual training is that of \citet{BERT} or \citet{conneau2019pretraining}, who train a masked language model (M-BERT, XLM) on a concatenation of monolingual corpora in different languages to learn a joint structure of these languages together. While several authors \citep{pires2019,wu-dredze-2019-beto, crossbert, libovicky2019} bring evidence of cross-lingual transfer within the model, its internal representations are not entirely language agnostic. 

We propose a method to further align representations from such models into the cross-lingual space and use them to derive sentence embeddings. Our approach is completely unsupervised and is applicable even for very distant language pairs. The proposed method outperforms previous unsupervised approaches on the BUCC 2018\footnote{11th Workshop on Building and Using Comparable Corpora} shared task, and is even competitive with several supervised baselines.
%We use the embeddings to mine parallel sentences from comparable corpora.
%that pretrained multilingual models exhibit a surprisingly strong ability to transfer knowledge from one language to another

%A popular approach to solving natural language processing tasks is to use a language representation model such as BERT \citep{BERT} or RoBERTa \citep{roberta} pretrained on large unlabeled data. Recent work \citep{BERT,conneau2019pretraining} has focused on the possibility of such pretraining in multilingual context on texts in different languages (e.g. M-BERT, XLM) and showed that the model is able to learn a joint structure of these languages together. 

%This work uses internal representations from a pretrained multilingual model to derive language agnostic sentence embeddings and use them to mine parallel sentences from comparable corpora. Our approach is completely unsupervised and is thus applicable even for very low-resource language pairs. It offers a method to mine parallel sentences even if there are no parallel resources to begin with. 
 %, using only monolingual data for training.

The paper is organized as follows. \cref{sec:related} gives an overview of related work; \cref{sec:proposed} introduces the proposed method; \cref{sec:results} describes the experiments and reports the results. \cref{sec:concl} concludes.
%which is problematic especially in the context of parallel corpus mining, where we would greatly benefit from a method applicable 

%The proposed method outperforms previous unsupervised approaches in BUCC, and is even competitive with several supervised baselines.

%Other applications of multilingual sentence embeddings include corpus filtering, or semantic similarity search.

% The advantage of our method is that it is only based on monolingual data and therefore can be used for low-resource languages. %A further advantage arises from the fact that fixed-length embeddings can be efficiently used for search related tasks as they do not require joint encoding of every sentence pair combination. % which is quadratic in the number of sentences

% i.e. identifying parallel sentences in non-aligned texts,
%is evaluated in our experiments
%Our experiments show results of a corpus filtering task (BUCC). 
\section{Related Work}
\label{sec:related}

Related research comprises supervised methods to model multilingual sentence embeddings and unsupervised methods to model multilingual word embeddings which can be aggregated into sentences. Furthermore, our approach is closely related to the recent research in cross-lingual language model (LM) pretraining.

%Unsupervised sentence embeddings have not yet been sufficiently explored. 
%The first attempts to derive unsupervised sentence embeddings 

\textbf{Supervised multilingual sentence embeddings.} 
The state-of-the-art performance in parallel data mining is achieved by LASER \citep{Artetxe2019laser} -- a multilingual BiLSTM model sharing a single encoder for 93 languages trained on parallel corpora to produce language agnostic sentence representations.
Similarly, \citet{schwenk-douze-2017-learning,schwenk-2018-filtering, Espana_Bonet_2017} derive sentence embeddings from internal representations of a neural machine translation system with a shared encoder. The universal sentence encoder (USE) \citep{cer-etal-2018-universal, yang2019multilingual} family covers sentence embedding models with a multi-task dual-encoder training framework including the tasks of question-answer prediction or natural language inference. \citet{guo-etal-2018-effective} directly optimize the cosine similarity between the source and target sentences using a bidirectional dual-encoder. %\citet{Yang_2019} enhance the model with an \textit{additive margin softmax} loss to separate translations from nearby non-translations. %They also train a separate classifier to rescore candidate sentences and retrieve final sentence pairs.
These approaches rely on heavy supervision by parallel corpora which is not available for low-resource languages.

\textbf{Unsupervised multilingual word embeddings.} 
Cross-lingual embeddings of words can be obtained by post-hoc alignment of monolingual word embeddings \citep{mikolov2013mapping} and mean-pooled with IDF weights to represent sentences \citep{litschko2019}. Unsupervised techniques to find a linear mapping between embedding spaces were proposed by \citet{artetxe2018vecmap} and \citet{conneau2017muse}, using iterative self-learning or adversarial training. Several recent studies \citep{patra-etal-2019-bilingual, ormazabal} criticize this simplified approach, showing that even the embedding spaces of closely related languages are not isometric. \citet{vulic-etal-2019-really} question the robustness of unsupervised mapping methods in challenging circumstances. %However, since the mapping is applied to static (non-contextualized) embeddings, these approaches give up on the contextual information which could be exploited in the sentence embeddings construction.  %\citet{litschko2019} aggregate aligned word embedings by an IDF weighted average and show the superiority of this method over a direct query translation for cross-lingual information retrieval.

\textbf{Cross-lingual LM pretraining.} 
\citet{ma2019universal,sentBERT} derive monolingual sentence embeddings by mean-pooling contextualized word embeddings from BERT. \citet{schuster_2019, wang-etal-2019-cross} propose mapping such contextualized embeddings into the multilingual space and report favorable results on the task of dependency parsing.  \citet{pires2019} extract contextualized embeddings directly from unsupervised multilingual LMs and use them for parallel sentence retrieval. Other authors improve the alignment of representations in a multilingual LM using a parallel corpus as an anchor \citep{Cao2020Multilingual} or using iterative self-learning \citep{wang2019mono}. None of these works apply multilingual embeddings to mine parallel sentences. Our work is the first in improving unsupervised cross-lingual models using additional unsupervised information. 

%\citep{BERT,conneau2019pretraining,wada-etal-2019-unsupervised}

\section{Proposed Method}
\label{sec:proposed}

We propose a method to enhance the cross-lingual ability of a pretrained multilingual model by fine-tuning it on a small synthetic parallel corpus. The parallel corpus is obtained via unsupervised machine translation (MT) so the method remains unsupervised. In this section, we describe the pretrained model (\cref{ssec:pretr}), the fine-tuning objective (\cref{ssec:tlm}) and the extraction of sentence embeddings (\cref{ssec:se}). We provide details on the unsupervised MT system in \cref{ssec:umt}. %We use its contextualized embeddings to derive fixed-length sentence representations and evaluate how language agnostic they are.

\subsection{XLM Pretraining}
\label{ssec:pretr}
%We use the \texttt{XLM-100}\footnote{\url{https://github.com/facebookresearch/XLM}} pretrained model with 16 layers, 16 attention heads and a hidden unit size of 1280. The model was trained using the masked language model (MLM) objective on Wikipedia texts in 100 languages split into streams of 256 tokens. The input text is processed in BPE subword units \citep{sennrich} and the vocabulary is shared for all languages. Its vocabulary is of size 200k.

The starting point for our experiments is a cross-lingual language model (XLM) \citep{conneau2019pretraining} of the BERT family pretrained on concatenated monolingual texts in 100 languages using the masked language model (MLM) training objective \citep{BERT}. The model processes the input in BPE subword units \citep{sennrich} with a shared vocabulary for all languages.  In this work, we use the publicly available pretrained model XLM-100\footnote{\url{https://github.com/facebookresearch/XLM}} \citep{conneau2019pretraining} with 16 transformer layers, 16 attention heads and a hidden unit size of 1280. The model was trained on monolingual corpora in 100 languages with 
the BPE vocabulary of 240k subwords.

%We use the XLM \citep{conneau2019pretraining} multilingual LM which was pretrained on concatenated texts in 100 languages using the MLM objective \citep{BERT}. The input text is processed in BPE subword units \citep{sennrich} and the vocabulary is shared for all languages. We translate a small amount of monolingual texts (20k sentences) by an unsupervised MT system to create a synthetic parallel corpus and use it to fine-tune the XLM model with a translation objective. Finally, we derive sentence embeddings by aggregating the per-word outputs of the internal layers of the model. 

\subsection{XLM Fine-tuning with a Translation Objective}
\label{ssec:tlm}
When parallel data is available, it can be leveraged in training of the multilingual language model using a translation language model loss (TLM) \citep{conneau2019pretraining}. Pairs of sentences are concatenated, random tokens are masked from both sentences and the model is trained to fill in the blanks by attending to any of the words of the two sentences. The Transformer self-attention layers thus have the capacity to enrich word representations with the information about their monolingual context as well as their translation counterparts. This explicit cross-lingual training objective further enhances the alignment of the embeddings in the cross-lingual space.

We use this objective to fine-tune the pretrained model on a small synthetic parallel data set obtained via unsupervised MT for one language pair, aiming to improve the overall cross-lingual alignment of the internal representations of the model. In our experiments, we also compare the performance to fine-tuning on small authentic parallel corpora.

\subsection{Sentence Embeddings}
\label{ssec:se}
Pretrained language models produce contextual representations capturing the semantic and syntactic properties of word (subword) tokens in their variable context \citep{BERT}. Contextualized embeddings can be derived from any of the internal layer outputs of the model. We tune the choice of the layer on the task of parallel sentence matching and conclude that the best cross-lingual performance is achieved at the 12th (5th-to-last) layer. Therefore, we use the representations from this layer in the rest of this paper. The evaluation across layers is summarized in \cref{fig:psm} in \cref{sec:analysis}. %We compare sentence embeddings derived from different layers and conclude that the best cross-lingual performance is consistently achieved at the 12th (5th-to-last) layer. Therefore, we use the representations from this layer in the rest of this paper.  

%Each subword of the vocabulary is tied to one input embedding vector. As the vector is passed through the encoder, it is enriched with information about its context, position etc. 

%Previous research suggests that different encoder layers represent different linguistic phenomena \citep{jawahar-etal-2019-bert} and achieve variable performance on downstream probing tasks \citep{Liu_2019}. Our experiments show that also the cross-lingual ability of the model changes across its layers. 

Aggregating subword embeddings to fixed-length sentence representations necessarily leads to an information loss. We compose sentence embeddings from subword representations by simple element-wise averaging. Even though mean-pooling is a naive approach to subword aggregation, it is often used for its simplicity \citep{sentBERT,ruiter-etal-2019-self,ma2019universal} and in our scenario it yields better results than max-pooling.

\subsection{Unsupervised Machine Translation}
\label{ssec:umt}

Our unsupervised MT model follows the approach of \citet{conneau2019pretraining}. It is a Transformer model with an encoder-decoder architecture. Both the encoder and the decoder are shared across languages and they are initialized with a pretrained bilingual LM to bootstrap the training.  Both the encoder and the decoder have 6 layers, 8 attention heads and a hidden unit size of 768. The system is trained using the unsupervised MT training pipeline of denoising and back-translation \citep{lample2018only}. %as proposed by \citep{lample2018only} with a de-noising and online back-translation. %on two sub-tasks, learning to recover a sentence from a noisy version of itself (denoising) and to recover a sentence from its synthetic translation (back-translation). The training pipeline consists of switching between these two sub-tasks, one batch each in every step.

\section{Experiments \& Results}
\label{sec:results}

\begin{table*}[t]
    \centering\small
    \begin{tabular}{l|c|c|c|c || c | c}
                 &  \textbf{en-de} & \textbf{en-fr} & \textbf{en-ru} & \textbf{en-zh}  &  \multicolumn{2}{c}{\textbf{Supervision}}  \\
         \hline
         \small{\citet{LEONG18.7}} & - & - & - & 56.00 & bitext & 0.5M sent. \\
        \small{\citet{bouamor2018}} & -  & 76.00 & - & - & bitext & 2M sent. \\
                \small{\citet{schwenk-2018-filtering}} & 76.90 &  75.80  &73.80 & 71.60 & 9-way parallel& 2M sent. \\

        \small{\citet{AZPEITIA18.6}} & 85.52 & 81.47  & 81.30& 77.45  &  bitext & 2-9M sent. \\
       %\small{\citet{Yang_2019}} & \textbf{97.24} & \textbf{96.96}  & \textnf{93.38} &  \textbf{96.00} &  web crawled bitext & 400M sent. \\ 
              
       \small{\citet{Artetxe2019laser}} & \textbf{ 96.19 } & \textbf{93.91} & \textbf{93.30} & \textbf{92.27} & 2- or 3-way parallel & 223M sent. \\
       \hline\hline
        \small{Unsup. baseline (Word Mapping)}  & 32.04 & 32.94  & 17.68  & 20.65   & none & n/a \\
        \small{Unsup. baseline (Vanilla XLM)*}  & 62.10 & 64.77 & 61.65 & 44.79 & none & n/a  \\

       %\hline % OB has hidden this line
       \small{\textbf{Proposed method* (en$\leftrightarrow$de)}} & \textbf{80.06} & \textbf{78.77} & \textbf{77.16} & \textbf{67.04} & none & 20k sent.** \\ %Synthetic (20k sentences) 
    \end{tabular}
    \caption{F1 score on the parallel sentence mining task (BUCC test set). The supervised (upper part) and unsupervised (lower part) winners are highlighted in bold. * The model was pretrained on Wikipedia. ** Synthetic translations produced by unsupervised MT.}
    \label{tab:bucc}
\end{table*}

\begin{table*}
    \centering\small
    \begin{tabular}{l|c|c|c|c | c | c | c}
                 &  \textbf{en-de} & \textbf{en-fr} & \textbf{en-ru} & \textbf{en-zh} & \textbf{en-kk}  & \textbf{cs-zh} & \textbf{de-ru} \\
         \hline
                 
       \small{\citet{Artetxe2019laser}} & \textbf{90.30} & \textbf{87.38} & \textbf{94.34} & \textbf{83.92} & 12.07 & \textbf{73.41} & \textbf{88.39}\\
       \hline\hline
        \small{Unsup. baseline (Word Mapping)}  & 28.45 & 30.79  & 17.81 &  16.04  & 2.28 & 10.86 & 19.55 \\
        \small{Unsup. baseline (Vanilla XLM)}  & 72.58  & 71.92 & 72.90 & 59.26 & 24.00 & 43.00 & 58.29  \\

       %\hline % OB has hidden
       \small{\textbf{Proposed method (en$\leftrightarrow$de)}} & \textbf{79.32 } & \textbf{77.05} & \textbf{80.98} & \textbf{65.49} & \textbf{35.41} & \textbf{48.79} & \textbf{65.91} \\ %SyntWe evaluate on two datasets -- the original BUCC datahetic (20k sentences) 
    \end{tabular}
    \caption{F1 score on the parallel sentence mining task (News test set). The supervised and unsupervised winners are highlighted in bold. \citet{Artetxe2019laser} values obtained using the public implementation of the LASER toolkit.}
    \label{tab:bucc-news}
\end{table*}

In this section, we empirically evaluate the quality of our cross-lingual sentence embeddings and compare it with state-of-the-art supervised methods and unsupervised baselines. We evaluate the proposed method on the task of parallel corpus mining and parallel sentence matching. We fine-tune two different models using English-German and Czech-German synthetic parallel data.

\subsection{Data}
The XLM model was pretrained on the Wikipedia corpus of 100 languages \citep{conneau2019pretraining}. The monolingual data for fine-tuning was sampled from NewsCrawl 2018 (10k Czech sentences, 10k German sentences, 10k English sentences). 

Monolingual training data for the unsupervised MT models was obtained from NewsCrawl 2007-2008 (5M sentences per language). The text was cleaned and tokenized using standard Moses \citep{moses} tools and segmented into BPE units based on 60k BPE splits.

\subsection{Experiment Details}
To generate synthetic data for fine-tuning, we train two unsupervised MT models (Czech-German, English-German) using the same method and parameters as in \citet{conneau2019pretraining} on 8 GPUs for 24 hours. We use these models to translate 10k sentences in each language. The translations are coupled with the originals into two parallel corpora of 20k synthetic sentence pairs.

The small synthetic parallel corpora obtained in the first step are used to fine-tune the pretrained XLM-100 model using the TLM objective. We measure the quality of induced cross-lingual embeddings from different layers on the task of parallel sentence matching described in \cref{ssec:PSR} and observe the best results at the 12th layer after fine-tuning for one epoch with a batch size of 8 sentences and all other pretraining parameters intact. The development accuracy decreases with fine-tuning on a larger data set.% or longer training. %Finally, we construct sentence embeddings as described in \cref{ssec:se}. %During fine-tuning, we decrease the learning rate to $10^{-5}$, set the batch size to 8 sentences per batch and keep all other pretraining parameters intact.

%When constructing sentence embeddings, we extract the representations of the input tokens from the 12th layer of the fine-tuned model and mean-pool them into fixed-length sentence embeddings with 1280 dimensions, masking the special tokens ([CLS], [SEP]). We experimented with other pooling strategies (max-pooling, [CLS] token) and concluded that mean-pooling yields the best results.

\subsection{Baselines}
We assess our method against two unsupervised baselines to separately measure the fine-tuning effect on the XLM model and to compare our results to another possible unsupervised approach based on post-hoc alignment of word embeddings.  % relying either on contextualized embeddings derived from a cross-lingual language model without any fine-tuning (Vanilla XLM) or static word embeddings resulting from a post-hoc alignment of Word2Vec embeddings (Word Mapping).

%We use the pretrained XLM-100 model without further fine-tuning and extract the 12th layer token representations mean-pooled into sentence embeddings.

\textbf{Vanilla XLM}. Contextualized token representations are extracted from the 12th layer of the original XLM-100\footnote{Using M-BERT model yielded similar results to XLM.} model and mean-pooled into sentence embeddings. 

\textbf{Word Mapping}. We use Word2Vec embeddings with 300 dimensions pretrained on NewsCrawl and map them into the cross-lingual space using the unsupervised version of VecMap \citep{artetxe2018vecmap}. As above, word embeddings are aggregated by mean-pooling to represent sentences.\footnote{Weighting word embeddings by their sentence frequency (IDF) did not lead to a significant improvement over a simple average.} % strong results so they are simply mean-pooled to represent sentences.
%
%  %However, mean-pooling static word embedding  %To represent a sentence, we make an average of the mapped static embedings of each word of the sentence.

\subsection{Evaluation I: Parallel Corpus Mining}
\label{ssec:BUCC}

%The motivation behind parallel corpus mining is to obtain parallel texts from comparable or non-aligned documents.

We measure the performance of our method on the BUCC shared task of parallel corpus mining where the system is expected to search two comparable non-aligned corpora and identify pairs of parallel sentences. We evaluate on two data sets -- the original BUCC 2018 corpus created by inserting parallel sentences into monolingual texts extracted from Wikipedia \citep{bucc} and a new BUCC-like data set (News train and test) which we created by shuffling 10k parallel sentence from News Commentary into 400k monolingual sentences from News Crawl. The BUCC and News data sets are comparable in size and contain parallel sentences from the same source, but differ in overall domain.

%Our evaluation is consistent with a scenario where one trains the model using the comparable corpora one wishes to mine. 
 
%The task is evaluated for German (414k sentences), French (272k sentences), Russian (461k sentences) and Chinese (95k sentences), each coupled with an English corpus of similar size. There are 2-3\% parallel sentences for each language pair. The BUCC corpora were created by inserting parallel sentences into monolingual texts extracted from Wikipedia \citep{bucc}. 

In order to score all candidate sentence pairs, we use the margin-based
approach of \citet{schwenkartexte2018} which was proved to eliminate the
hubness problem of embedding spaces and yield superior results
\citep{Artetxe2019laser}. The score relies on cosine similarity to measure the
distance between sentences but it is defined in relative terms to the average
cosine similarity between the two sentences and their nearest neighbors. The
optimal threshold for filtering the translation pairs is learned by tuning on
the train set F1 scores.  \cref{tab:bucc,tab:bucc-news} show the results of our proposed model on the BUCC and News test sets, resp., comparing them to related work and unsupervised baselines.
 
When comparing our method to related work, it must be noted that the XLM model
was pretrained on Wikipedia and therefore has seen the monolingual BUCC
sentences during training. This could result in an advantage over other
systems, as the model could exploit the fact that it has seen the non-parallel
part of the comparable corpus during training. However, since both the
proposed method an the \textit{vanilla XLM} baseline suffer from this, their
results remain comparable. We also report results on the News test set which
is free from such potential bias (\cref{tab:bucc-news}).  %

 %Including comparable corpora in the training data does not 
 
%Since the XLM model was pretrained on Wikipedia and therefore has seen the monolingual BUCC sentences during training, we cannot directly compare our model to other authors who did not use Wikipedia for training. To provide a full overview of related work on this task, we report their results in \cref{tab:bucc}.

\begin{table*}
    \centering\small
    \begin{tabular}{l|c|c|c|c|c|c|c|c}
         &   \textbf{de-en} & \textbf{cs-en} & \textbf{cs-de} & \textbf{cs-fr} & \textbf{cs-ru}  & \textbf{fr-es}  & \textbf{fr-ru} & \textbf{es-ru}  \\
         
         \hline
         
        \small{\citet{Artetxe2019laser}} & 98.78 & 99.08 & 99.23 & 99.37 & 98.77 & 99.42 & 98.60 & 98.77  \\
       \hline
       \hline
        \small{Unsup. baseline (Word Mapping)} & 60.60  & 55.03  &  75.35 &  43.33 & 79.87 & 71.07 & 41.25  & 53.87 \\
        \small{Unsup. baseline (Vanilla XLM)} & 87.15 & 79.83 & 82.87 & 80.55 & 85.15 & 91.07 & 85.28 & 85.73\\
       \hline
       \small{\textbf{Proposed method (en$\leftrightarrow$de) }} & 93.97 & \textbf{90.47} & 90.48 & \textbf{90.07} & 92.23 & \textbf{94.68} & \textbf{91.80} & \textbf{91.92}  \\
      \small{\textbf{Proposed method (cs$\leftrightarrow$de) }} & \textbf{94.43} & 90.15 & \textbf{90.50} & 89.48 & \textbf{92.33} & 94.65 & 91.72 & 91.25  \\

    \end{tabular}
    \caption{Accuracy on a parallel sentence matching task (\textit{newstest2012}) averaged over both matching directions.}
    \label{tab:news}
\end{table*}

The results reveal that TLM fine-tuning brings a substantial improvement over the initial pretrained model trained only using the MLM objective (\textit{vanilla XLM}). In terms of the F1 score, the gain across four BUCC language pairs is 14.0-22.3 points. Even though the fine-tuning focused on a single language pair (English-German), the improvement is notable for all evaluated language pairs. The largest margin of 21.6 points is observed for the English-Chinese mining task. We observe that using a small parallel data set of authentic translation pairs instead of synthetic ones does not have a significant effect. 

The weak results of the \textit{word mapping} baseline can be partially attributed to the superiority of contextualized embeddings for representation of sentences over static ones. Furthermore, word mapping relies on the questionable assumption of isomorphic embedding spaces which weakens its performance especially for distant languages. In our proposed model, it is possible that joint training of contextualized representations induces an embedding space with more convenient geometric properties which makes it more robust to language diversity. % As suggested by \citet{ormazabal}, separately trained embedding spaces suffer from 

Although the performance of our model generally lags far behind the supervised LASER benchmark, it is valuable because of its fully unsupervised nature and it works even for distant languages such as Chinese-Czech or English-Kazakh. 

%Our BUCC results even supersede some earlier supervised models such as \citet{schwenk-2018-filtering} or \citet{bouamor2018}. 
\begin{table*}
    \centering\small
\begin{tabular}{lcccccccccccccc}
\hline
 & \textbf{af} & \textbf{ar} & \textbf{az} & \textbf{be} & \textbf{bg} & \textbf{ca} & \textbf{cs} & \textbf{de} & \textbf{el} & \textbf{eo} & \textbf{et} & \textbf{fi} & \textbf{fy} & \textbf{gl} \\
\textbf{Sup. baseline} &     89.5 &    92.0 &    66.0 &    66.2 &    95.0 &    95.9 &    96.5 &    99.0 &    95.0 &    97.2 &    96.7 &    96.3 &    51.7 &    95.5 \\
\textbf{Vanilla XLM} &     38.1 &    19.9 &    25.1 &    33.7 &    36.2 &    51.0 &    31.5 &    65.0 &    27.0 &    45.8 &    19.8 &    31.4 &    37.0 &    51.4 \\
\multicolumn{15}{l}{\textbf{Proposed method:}} \\
\quad\textbf{en$\leftrightarrow$de} (synth) &     57.3 &    41.1 &    46.3 &    58.4 &    56.0 &    66.9 &    53.5 &    83.1 &    51.3 &    68.0 &    39.0 &    47.5 &    48.6 &    66.9 \\
\quad\textbf{cs$\leftrightarrow$de} (synth) &     54.2 &    41.2 &    44.2 &    61.8 &    60.7 &    68.9 &    59.9 &    87.3 &    53.1 &    67.4 &    41.4 &    49.5 &    44.8 &    67.3 \\
\quad\textbf{en$\leftrightarrow$kk} (auth) &     58.4 &    45.6 &    51.4 &    60.2 &    59.2 &    72.6 &    53.9 &    87.0 &    54.6 &    72.1 &    43.4 &    51.3 &    51.7 &    72.2 \\
\quad\textbf{en$\leftrightarrow$ne} (auth) &     59.9 &    46.6 &    54.2 &    63.1 &    62.9 &    71.0 &    57.6 &    85.0 &    51.0 &    71.2 &    44.6 &    52.7 &    48.6 &    71.0 \\
\hline
 & \textbf{hi} & \textbf{hr} & \textbf{ia} & \textbf{is} & \textbf{id} & \textbf{ja} & \textbf{ka} & \textbf{kk} & \textbf{ku} & \textbf{la} & \textbf{lt} & \textbf{mk} & \textbf{ml} & \textbf{mn} \\
\textbf{Sup. baseline} &     94.7 &    97.2 &    95.2 &    95.6 &    94.5 &    91.8 &    35.9 &    18.6 &    17.2 &    58.5 &    96.2 &    94.7 &    96.9 &     8.2 \\
\textbf{Vanilla XLM} &     26.2 &    47.2 &    57.3 &    25.0 &    46.4 &    29.5 &    22.1 &    17.4 &    10.6 &    15.5 &    22.0 &    25.8 &    17.4 &    12.6 \\
\multicolumn{15}{l}{\textbf{Proposed method:}} \\
\quad\textbf{en$\leftrightarrow$de} (synth) &     53.4 &    68.2 &    71.4 &    43.1 &    64.9 &    54.4 &    41.4 &    33.6 &    16.8 &    24.9 &    43.9 &    48.8 &    51.6 &    29.0 \\
\quad\textbf{cs$\leftrightarrow$de} (synth) &     51.7 &    71.8 &    70.5 &    43.7 &    64.1 &    53.3 &    39.8 &    34.7 &    16.2 &    27.7 &    46.2 &    51.1 &    44.3 &    24.5 \\
\quad\textbf{en$\leftrightarrow$kk} (auth) &     60.3 &    71.3 &    79.5 &    45.0 &    66.4 &    59.6 &    44.0 &    46.1 &    20.0 &    28.6 &    46.2 &    54.7 &    54.0 &    32.7 \\
\quad\textbf{en$\leftrightarrow$ne} (auth) &     59.3 &    72.1 &    75.7 &    47.1 &    67.8 &    59.6 &    47.8 &    38.4 &    20.9 &    30.0 &    47.7 &    53.8 &    56.0 &    34.9 \\
\hline
 & \textbf{mr} & \textbf{ms} & \textbf{ne} & \textbf{nn} & \textbf{oc} & \textbf{sl} & \textbf{sr} & \textbf{sv} & \textbf{ta} & \textbf{te} & \textbf{tl} & \textbf{uk} & \textbf{ur} & \textbf{yi} \\
\textbf{Sup. baseline} &     91.5 &    96.4 &    20.6 &    88.3 &    61.2 &    95.9 &    95.3 &    96.6 &    69.4 &    79.7 &    50.5 &    94.5 &    81.9 &     5.7 \\
\textbf{Vanilla XLM} &     15.3 &    52.0 &    21.3 &    49.9 &    20.0 &    34.7 &    35.9 &    47.2 &    11.9 &    14.1 &    14.6 &    38.0 &    19.3 &     9.9 \\
\multicolumn{15}{l}{\textbf{Proposed method:}} \\
\quad\textbf{en$\leftrightarrow$de} (synth) &     37.3 &    67.0 &    32.8 &    66.8 &    34.3 &    54.9 &    58.6 &    69.7 &    40.9 &    44.7 &    24.0 &    66.1 &    43.7 &    22.1 \\
\quad\textbf{cs$\leftrightarrow$de} (synth) &     34.2 &    65.4 &    31.4 &    67.5 &    35.9 &    59.2 &    64.8 &    71.8 &    31.9 &    37.8 &    20.4 &    70.4 &    43.8 &    22.8 \\
\quad\textbf{en$\leftrightarrow$kk} (auth) &     41.9 &    69.8 &    37.3 &    69.2 &    40.3 &    58.0 &    64.3 &    73.3 &    42.8 &    44.0 &    24.4 &    71.6 &    48.2 &    25.8 \\
\quad\textbf{en$\leftrightarrow$ne} (auth) &     43.5 &    72.1 &    42.8 &    69.2 &    36.9 &    58.8 &    65.0 &    72.0 &    41.7 &    53.2 &    26.8 &    71.0 &    49.9 &    26.7 \\
\hline
\end{tabular}  
\caption{Accuracy on a parallel sentence matching task (\textit{Tatoeba}) averaged over both matching directions (to and from English). The supervised baseline was obtained using the public implementation of the LASER model \citep{Artetxe2019laser}. Our proposed models were fine-tuned on synthetic parallel data (en$\leftrightarrow$de, cs$\leftrightarrow$de) and authentic parallel data (en$\leftrightarrow$kk, en$\leftrightarrow$ne). } 

    \label{tab:tatoeba}
\end{table*}

\subsection{Evaluation II: Parallel Sentence Matching}
\label{ssec:PSR}

%XXX Mention named entities problem for vecmap

To assess the effect of proposed fine-tuning on other language pairs not covered by BUCC, we evaluate our embeddings on the task of parallel sentence matching (PSM). The task entails searching a pool of shuffled parallel sentences to recover correct translation pairs.  Cosine similarity is used for the nearest neighbor search.

We first evaluate the pairwise matching accuracy on a \textit{newstest} multi-way parallel data set of 3k sentences in 6 languages.\footnote{Czech, English, French, German, Russian, Spanish} We use \textit{newstest2012} for development and \textit{newstest2013} for testing. The results in \cref{tab:news} show that the fine-tuned model is able to match correct translations in 90-95\% of cases, depending on the language pair, which is $\sim$7\% more than \textit{vanilla XLM}. It is notable that the model which was only fine-tuned on English-German synthetic parallel data has a positive effect on completely unrelated language pairs as well (e.g. Russian-Spanish, Czech-French). 

Since the greatest appeal of parallel corpus mining is to enhance the resources for low-resource languages, we also measure the PSM accuracy on the Tatoeba \citep{Artetxe2019laser} data set of 0.5--1k sentences in over 100 languages aligned with English. Aside from the two completely unsupervised models, we fine-tune two more models on small authentic parallel data in English-Nepali (5k sentence pairs from the Flores development sets) and English-Kazakh (10k sentence pairs from News Commentary). \Cref{tab:tatoeba} confirms that the improvement over \textit{vanilla XLM} is present for every language we evaluated, regardless on the language pair used for fine-tuning. Although we initially hypothesized that the performance of the English-German model on English-aligned language pairs would exceed the German-Czech model, their results are equal on average. Fine-tuning on small authentic corpora in low-resource languages exceeds both by a slight margin.%Using small authentic corpora in low-resource languages yielded on average higher accuracy than using synthetic corpora in high-resource languages. 

The results are clearly sensitive to the amount of monolingual sentences in the Wikipedia corpus used for XLM pretraining and the matching accuracy of very low-resource languages is significantly lower than we observed for high-resource languages. However, the benefits of fine-tuning are substantial (around 20 percentage points) and for some languages the results even reach the supervised baseline (e.g. Kazakh, Georgian, Nepali). %We observe low matching accuracies for very low-resource languages %Although the matching accuracy is significantly lower than for high-resource languages, the sentence representations ar

It seems that explicitly aligning one language pair during fine-tuning propagates through the shared parameters and improves the overall representation alignment, making the contextualized embeddings more language agnostic. The propagation effect could also positively influence the ability of cross-lingual transfer within the model in downstream tasks. A verification of this is left to future work.  % Representations of languages with small Wikipedias ($\sim$100k sentences, e.g. Nepali, Khmer) are not aligned well enough to perform filtering using \textit{vanilla XLM} or our method.
 
%We observe that fine-tuning the model for longer or using more synthetic sentences leads to diverging results. 

\begin{figure}[h!]
    \centering
    \includegraphics[width=\columnwidth]{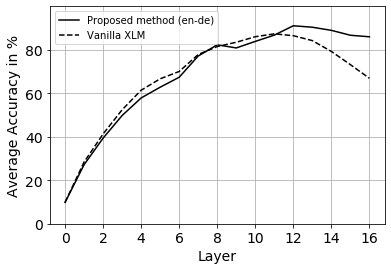}
    \caption{Average PSM accuracy on \textit{newstest2012} before and after fine-tuning from the input embedding layer (0th) to the deepest layer (16th).}
    \label{fig:psm}
\end{figure}

\subsection{Analysis: Representations Across Layers}
\label{sec:analysis}
We derive sentence embeddings from all layers of the model and show PSM results on the development set averaged over all language pairs in \cref{fig:psm}, both before and after fine-tuning. The accuracy differs substantially across the model depth, the best cross-lingual performance is consistently achieved around the 12th (5th-to-last) layer of the model. The TLM fine-tuning affects especially the deepest layers.

\section{Conclusion}
\label{sec:concl}

We proposed a completely unsupervised method to train multilingual sentence embeddings which can be used for building a parallel corpus with no previous translation knowledge. 

 We show that fine-tuning an unsupervised multilingual model with a translation objective using as little as 20k synthetic translation pairs can significantly enhance the cross-lingual alignment of its representations. Since the synthetic translations were obtained from an unsupervised MT system, the entire procedure requires no authentic parallel sentences for training.  
 
 %We evaluate our sentence embeddings on the tasks of parallel data mining and parallel sentence matching and report a significant improvement over the original XLM model which we set as our baseline.  % Our method is even competitive with several supervised benchmarks.
 
 Our sentence embeddings yield significantly better results on the tasks of parallel data mining and parallel sentence matching than our unsupervised baselines.  Interestingly, targeting only one language pair during the fine-tuning phase suffices to propagate the alignment improvement to unrelated languages. It is therefore not necessary to build a working MT system for every language pair we wish to mine. 
 
 The average F1 margin across four language pairs on the BUCC task is $\sim$17 points over the original XLM model and $\sim$7 on the News dataset where only one of the evaluated language pairs was seen during fine-tuning.
 The gain in accuracy in parallel sentence matching across 8 language pairs is 7.2\% absolute, lagging only 7.1\% absolute behind supervised methods.

% Furthermore, fine-tuning the model with the translation objective for one language pair suffices to propagate the alignment improvement to other languages as well.
 
%An interesting finding is that 
%Although we target only one language pair during the fine-tuning phase, it suffices to propagate the alignment improvement to other languages as well.

%Our embeddings can be used for efficient parallel data mining by scoring translation candidate pairs based on their cosine similarity.

%We show that sentence embeddings derived directly from unsupervised pretrained models (e.g. M-BERT) already contain some cross-lingual knowledge. Fine-tuning the model using as little as 10k synthetic translation pairs can further enhance the cross-lingual alignment. Furthermore, supervised fine-tuning on one language propagates the alignment improvement to other languages as well. 
For the future we would like to apply our model on other cross-lingual NLP tasks such as XNLI or cross-lingual semantic textual similarity.

\section*{Acknowledgments}

This study was supported in parts by the grants
SVV~260~575, 1050119 of the Charles University Grant Agency and
19-26934X of the Czech Science Foundation, by a Facebook Fellowship, the Basque Government excellence research group (IT1343-19), the Spanish MINECO (UnsupMT TIN2017‐91692‐EXP MCIU/AEI/FEDER, UE) and Project BigKnowledge (Ayudas Fundación BBVA a equipos de investigación científica 2018).

%This work has been using language resources and tools stored and distributed by the LINDAT/CLARIN project of the Ministry of Education, Youth and Sports of the Czech Republic (LM2015071).

%TODO
%map embeddings for vecmap (en-kk atd)
%try idf
%tlm - just mention
%pro co se da jeste pouzit multi embeddingy
%laser
%update train to test
%distant languages - check Vulic
%mention bert? 768 dims
%expand on small languages
%we might not need a working unsup MT system but rather finetune on more monolingual data to improve the representations
%german does not work for tlm
%add to comparison
\bibliography{biblio}

\begin{thebibliography}{35}
\expandafter\ifx\csname natexlab\endcsname\relax\def\natexlab#1{#1}\fi

\bibitem[{Artetxe et~al.(2018)Artetxe, Labaka, and Agirre}]{artetxe2018vecmap}
Mikel Artetxe, Gorka Labaka, and Eneko Agirre. 2018.
\newblock \href {https://www.aclweb.org/anthology/P18-1073.pdf} {A robust
  self-learning method for fully unsupervised cross-lingual mappings of word
  embeddings}.
\newblock In \emph{Proceedings of the 56th Annual Meeting of the ACL}, pages
  789--798, Melbourne. Association for Computational Linguistics.

\bibitem[{Artetxe and Schwenk(2019{\natexlab{a}})}]{schwenkartexte2018}
Mikel Artetxe and Holger Schwenk. 2019{\natexlab{a}}.
\newblock \href {https://doi.org/10.18653/v1/p19-1309} {Margin-based parallel
  corpus mining with multilingual sentence embeddings}.
\newblock \emph{Proceedings of the 57th Annual Meeting of the Association for
  Computational Linguistics}.

\bibitem[{Artetxe and Schwenk(2019{\natexlab{b}})}]{Artetxe2019laser}
Mikel Artetxe and Holger Schwenk. 2019{\natexlab{b}}.
\newblock \href {https://doi.org/10.1162/tacl_a_00288} {Massively multilingual
  sentence embeddings for zero-shot cross-lingual transfer and beyond}.
\newblock \emph{Transactions of the Association for Computational Linguistics},
  7:597–610.

\bibitem[{Azpeitia et~al.(2018)Azpeitia, Etchegoyhen, and
  Garcia}]{AZPEITIA18.6}
Andoni Azpeitia, Thierry Etchegoyhen, and Eva~Martínez Garcia. 2018.
\newblock \href {http://lrec-conf.org/workshops/lrec2018/W8/pdf/6_W8.pdf}
  {Extracting parallel sentences from comparable corpora with stacc variants}.
\newblock In \emph{Proceedings of the Eleventh International Conference on
  Language Resources and Evaluation (LREC 2018)}, Paris, France. European
  Language Resources Association (ELRA).

\bibitem[{Bouamor and Sajjad(2018)}]{bouamor2018}
Houda Bouamor and Hassan Sajjad. 2018.
\newblock \href
  {http://alt.qcri.org/~hsajjad/publications/papers/2018_BUCC_Bouamor_Comparable_Corpora.pdf}
  {H2@bucc18: Parallel sentence extraction from comparable corpora using
  multilingual sentence embeddings}.
\newblock In \emph{Proceedings of the Eleventh International Conference on
  Language Resources and Evaluation (LREC 2018)}, Paris, France. European
  Language Resources Association (ELRA).

\bibitem[{Cao et~al.(2020)Cao, Kitaev, and Klein}]{Cao2020Multilingual}
Steven Cao, Nikita Kitaev, and Dan Klein. 2020.
\newblock \href {https://openreview.net/forum?id=r1xCMyBtPS} {Multilingual
  alignment of contextual word representations}.
\newblock In \emph{International Conference on Learning Representations}.

\bibitem[{Cer et~al.(2018)Cer, Yang, Kong, Hua, Limtiaco, St.~John, Constant,
  Guajardo-Cespedes, Yuan, Tar, Strope, and Kurzweil}]{cer-etal-2018-universal}
Daniel Cer, Yinfei Yang, Sheng-yi Kong, Nan Hua, Nicole Limtiaco, Rhomni
  St.~John, Noah Constant, Mario Guajardo-Cespedes, Steve Yuan, Chris Tar,
  Brian Strope, and Ray Kurzweil. 2018.
\newblock \href {https://doi.org/10.18653/v1/D18-2029} {Universal sentence
  encoder for {E}nglish}.
\newblock In \emph{Proceedings of the 2018 Conference on Empirical Methods in
  Natural Language Processing: System Demonstrations}, pages 169--174,
  Brussels, Belgium. Association for Computational Linguistics.

\bibitem[{Conneau et~al.(2017)Conneau, Lample, Ranzato, Denoyer, and
  J{\'e}gou}]{conneau2017muse}
Alexis Conneau, Guillaume Lample, Marc'Aurelio Ranzato, Ludovic Denoyer, and
  Herv{\'e} J{\'e}gou. 2017.
\newblock \href {https://arxiv.org/pdf/1710.04087} {Word translation without
  parallel data}.
\newblock \emph{arXiv [e-Print archive]}.

\bibitem[{Devlin et~al.(2018)Devlin, Chang, Lee, and Toutanova}]{BERT}
Jacob Devlin, Ming{-}Wei Chang, Kenton Lee, and Kristina Toutanova. 2018.
\newblock \href {http://arxiv.org/abs/1810.04805} {{BERT:} pre-training of deep
  bidirectional transformers for language understanding}.
\newblock \emph{arXiv [e-Print archive]}, abs/1810.04805.

\bibitem[{Espana-Bonet et~al.(2017)Espana-Bonet, Varga, Barron-Cedeno, and van
  Genabith}]{Espana_Bonet_2017}
Cristina Espana-Bonet, Adam~Csaba Varga, Alberto Barron-Cedeno, and Josef van
  Genabith. 2017.
\newblock \href {https://doi.org/10.1109/jstsp.2017.2764273} {An empirical
  analysis of nmt-derived interlingual embeddings and their use in parallel
  sentence identification}.
\newblock \emph{IEEE Journal of Selected Topics in Signal Processing},
  11(8):1340–1350.

\bibitem[{Guo et~al.(2018)Guo, Shen, Yang, Ge, Cer, Hernandez~Abrego, Stevens,
  Constant, Sung, Strope, and Kurzweil}]{guo-etal-2018-effective}
Mandy Guo, Qinlan Shen, Yinfei Yang, Heming Ge, Daniel Cer, Gustavo
  Hernandez~Abrego, Keith Stevens, Noah Constant, Yun-Hsuan Sung, Brian Strope,
  and Ray Kurzweil. 2018.
\newblock \href {https://doi.org/10.18653/v1/W18-6317} {Effective parallel
  corpus mining using bilingual sentence embeddings}.
\newblock In \emph{Proceedings of the Third Conference on Machine Translation:
  Research Papers}, pages 165--176, Brussels, Belgium. Association for
  Computational Linguistics.

\bibitem[{Karthikeyan et~al.(2019)Karthikeyan, Wang, Mayhew, and
  Roth}]{crossbert}
Kaliyaperumal Karthikeyan, Zihan Wang, Stephen Mayhew, and Dan Roth. 2019.
\newblock \href {https://arxiv.org/abs/1912.07840} {Cross-lingual ability of
  multilingual {BERT}: An empirical study}.
\newblock \emph{arXiv [e-Print archive]}, abs/1912.07840.

\bibitem[{Koehn et~al.(2007)Koehn, Hoang, Birch, Callison-Burch, Federico,
  Bertoldi, Cowan, Shen, Moran, Zens, Dyer, Bojar, Constantin, and
  Herbst}]{moses}
Philipp Koehn, Hieu Hoang, Alexandra Birch, Chris Callison-Burch, Marcello
  Federico, Nicola Bertoldi, Brooke Cowan, Wade Shen, Christine Moran, Richard
  Zens, Chris Dyer, Ondrej Bojar, Alexandra Constantin, and Evan Herbst. 2007.
\newblock \href {https://www.aclweb.org/anthology/P07-2045} {{M}oses: Open
  source toolkit for statistical machine translation}.
\newblock In \emph{Proceedings of the 45th Annual Meeting of the ACL}, pages
  177--180, Prague. Association for Computational Linguistics.

\bibitem[{Lample and Conneau(2019)}]{conneau2019pretraining}
Guillaume Lample and Alexis Conneau. 2019.
\newblock \href {http://arxiv.org/abs/1901.07291} {Cross-lingual language model
  pretraining}.
\newblock \emph{arXiv [e-Print archive]}, abs/1901.07291.

\bibitem[{Lample et~al.(2018)Lample, Denoyer, and Ranzato}]{lample2018only}
Guillaume Lample, Ludovic Denoyer, and Marc'Aurelio Ranzato. 2018.
\newblock \href {http://arxiv.org/abs/1711.00043} {Unsupervised machine
  translation using monolingual corpora only}.
\newblock In \emph{Proceedings of the 6th International Conference on Learning
  Representations}.

\bibitem[{Leong et~al.(2018)Leong, Wong, and Chao}]{LEONG18.7}
Chongman Leong, Derek~F. Wong, and Lidia~S. Chao. 2018.
\newblock \href {http://lrec-conf.org/workshops/lrec2018/W8/pdf/7_W8.pdf}
  {Um-paligner: Neural network-based parallel sentence identification model}.
\newblock In \emph{Proceedings of the Eleventh International Conference on
  Language Resources and Evaluation (LREC 2018)}, Paris, France. European
  Language Resources Association (ELRA).

\bibitem[{Libovick\'{y} et~al.(2019)Libovick\'{y}, Rosa, and
  Fraser}]{libovicky2019}
Jind\v{r}ich Libovick\'{y}, Rudolf Rosa, and Alexander~M. Fraser. 2019.
\newblock \href {https://arxiv.org/abs/1911.03310} {How language-neutral is
  multilingual {BERT}?}
\newblock \emph{arXiv [e-Print archive]}, abs/1911.03310.

\bibitem[{Litschko et~al.(2019)Litschko, Glava\v{s}, Vulic, and
  Dietz}]{litschko2019}
Robert Litschko, Goran Glava\v{s}, Ivan Vulic, and Laura Dietz. 2019.
\newblock \href {https://doi.org/10.1145/3331184.3331324} {Evaluating
  resource-lean cross-lingual embedding models in unsupervised retrieval}.
\newblock In \emph{Proceedings of the 42nd International ACM SIGIR Conference
  on Research and Development in Information Retrieval}, SIGIR’19, page
  1109–1112, New York, NY, USA. Association for Computing Machinery.

\bibitem[{Ma et~al.(2019)Ma, Wang, Ng, Nallapati, and Xiang}]{ma2019universal}
Xiaofei Ma, Zhiguo Wang, Patrick Ng, Ramesh Nallapati, and Bing Xiang. 2019.
\newblock \href {https://arxiv.org/abs/1910.07973} {Universal text
  representation from bert: An empirical study}.
\newblock \emph{arXiv [e-Print archive]}, abs/1910.07973.

\bibitem[{Mikolov et~al.(2013)Mikolov, Le, and Sutskever}]{mikolov2013mapping}
Tomas Mikolov, Quoc~V. Le, and Ilya Sutskever. 2013.
\newblock \href {http://arxiv.org/abs/1309.4168} {Exploiting similarities among
  languages for machine translation}.
\newblock \emph{CoRR}, abs/1309.4168.

\bibitem[{Ormazabal et~al.(2019)Ormazabal, Artetxe, Labaka, Soroa, and
  Agirre}]{ormazabal}
Aitor Ormazabal, Mikel Artetxe, Gorka Labaka, Aitor Soroa, and Eneko Agirre.
  2019.
\newblock \href {https://www.aclweb.org/anthology/P19-1492} {Analyzing the
  limitations of cross-lingual word embedding mappings}.
\newblock In \emph{Proceedings of the 57th Annual Meeting of the ACL}, pages
  4990--4995, Florence. Association for Computational Linguistics.

\bibitem[{Patra et~al.(2019)Patra, Moniz, Garg, Gormley, and
  Neubig}]{patra-etal-2019-bilingual}
Barun Patra, Joel Ruben~Antony Moniz, Sarthak Garg, Matthew~R. Gormley, and
  Graham Neubig. 2019.
\newblock \href {https://doi.org/10.18653/v1/P19-1018} {Bilingual lexicon
  induction with semi-supervision in non-isometric embedding spaces}.
\newblock In \emph{Proceedings of the 57th Annual Meeting of the Association
  for Computational Linguistics}, pages 184--193, Florence, Italy. Association
  for Computational Linguistics.

\bibitem[{Pires et~al.(2019)Pires, Schlinger, and Garrette}]{pires2019}
Telmo Pires, Eva Schlinger, and Dan Garrette. 2019.
\newblock \href {https://doi.org/10.18653/v1/P19-1493} {How multilingual is
  multilingual {BERT}?}
\newblock In \emph{Proceedings of the 57th Annual Meeting of the ACL}, pages
  4996--5001, Florence. Association for Computational Linguistics.

\bibitem[{Reimers and Gurevych(2019)}]{sentBERT}
Nils Reimers and Iryna Gurevych. 2019.
\newblock \href {https://doi.org/10.18653/v1/d19-1410} {Sentence-bert: Sentence
  embeddings using siamese bert-networks}.
\newblock \emph{Proceedings of the 2019 Conference on Empirical Methods in
  Natural Language Processing and the 9th International Joint Conference on
  Natural Language Processing (EMNLP-IJCNLP)}.

\bibitem[{Ruiter et~al.(2019)Ruiter, Espa{\~n}a-Bonet, and van
  Genabith}]{ruiter-etal-2019-self}
Dana Ruiter, Cristina Espa{\~n}a-Bonet, and Josef van Genabith. 2019.
\newblock \href {https://doi.org/10.18653/v1/P19-1178} {Self-supervised neural
  machine translation}.
\newblock In \emph{Proceedings of the 57th Annual Meeting of the Association
  for Computational Linguistics}, pages 1828--1834, Florence, Italy.
  Association for Computational Linguistics.

\bibitem[{Schuster et~al.(2019)Schuster, Ram, Barzilay, and
  Globerson}]{schuster_2019}
Tal Schuster, Ori Ram, Regina Barzilay, and Amir Globerson. 2019.
\newblock \href {https://doi.org/10.18653/v1/n19-1162} {Cross-lingual alignment
  of contextual word embeddings, with applications to zero-shot dependency
  parsing}.
\newblock \emph{Proceedings of the 2019 Conference of the North}.

\bibitem[{Schwenk(2018)}]{schwenk-2018-filtering}
Holger Schwenk. 2018.
\newblock \href {https://doi.org/10.18653/v1/P18-2037} {Filtering and mining
  parallel data in a joint multilingual space}.
\newblock In \emph{Proceedings of the 56th Annual Meeting of the Association
  for Computational Linguistics (Volume 2: Short Papers)}, pages 228--234,
  Melbourne, Australia. Association for Computational Linguistics.

\bibitem[{Schwenk and Douze(2017)}]{schwenk-douze-2017-learning}
Holger Schwenk and Matthijs Douze. 2017.
\newblock \href {https://doi.org/10.18653/v1/W17-2619} {Learning joint
  multilingual sentence representations with neural machine translation}.
\newblock In \emph{Proceedings of the 2nd Workshop on Representation Learning
  for {NLP}}, pages 157--167, Vancouver, Canada. Association for Computational
  Linguistics.

\bibitem[{Sennrich et~al.(2016)Sennrich, Haddow, and Birch}]{sennrich}
Rico Sennrich, Barry Haddow, and Alexandra Birch. 2016.
\newblock \href {https://doi.org/10.18653/v1/P16-1162} {Neural machine
  translation of rare words with subword units}.
\newblock In \emph{Proceedings of the 54th Annual Meeting of the ACL (Volume 1:
  Long Papers)}, pages 1715--1725, Berlin. Association for Computational
  Linguistics.

\bibitem[{Vuli{\'c} et~al.(2019)Vuli{\'c}, Glava{\v{s}}, Reichart, and
  Korhonen}]{vulic-etal-2019-really}
Ivan Vuli{\'c}, Goran Glava{\v{s}}, Roi Reichart, and Anna Korhonen. 2019.
\newblock \href {https://doi.org/10.18653/v1/D19-1449} {Do we really need fully
  unsupervised cross-lingual embeddings?}
\newblock In \emph{Proceedings of the 2019 Conference on Empirical Methods in
  Natural Language Processing and the 9th International Joint Conference on
  Natural Language Processing (EMNLP-IJCNLP)}, pages 4407--4418, Hong Kong,
  China. Association for Computational Linguistics.

\bibitem[{Wang et~al.(2019{\natexlab{a}})Wang, Hou, Li, Tong, and
  Jiang}]{wang2019mono}
Shuai Wang, Lei Hou, Juanzi Li, Meihan Tong, and Jiabo Jiang.
  2019{\natexlab{a}}.
\newblock \href {https://doi.org/10.1007/978-3-030-32381-3_28} {\emph{Learning
  Multilingual Sentence Embeddings from Monolingual Corpus}}, pages 346--357.

\bibitem[{Wang et~al.(2019{\natexlab{b}})Wang, Che, Guo, Liu, and
  Liu}]{wang-etal-2019-cross}
Yuxuan Wang, Wanxiang Che, Jiang Guo, Yijia Liu, and Ting Liu.
  2019{\natexlab{b}}.
\newblock \href {https://doi.org/10.18653/v1/D19-1575} {Cross-lingual {BERT}
  transformation for zero-shot dependency parsing}.
\newblock In \emph{Proceedings of the 2019 Conference on Empirical Methods in
  Natural Language Processing and the 9th International Joint Conference on
  Natural Language Processing (EMNLP-IJCNLP)}, pages 5720--5726, Hong Kong,
  China. Association for Computational Linguistics.

\bibitem[{Wu and Dredze(2019)}]{wu-dredze-2019-beto}
Shijie Wu and Mark Dredze. 2019.
\newblock \href {https://doi.org/10.18653/v1/D19-1077} {Beto, bentz, becas: The
  surprising cross-lingual effectiveness of {BERT}}.
\newblock In \emph{Proceedings of the 2019 Conference on Empirical Methods in
  Natural Language Processing and the 9th International Joint Conference on
  Natural Language Processing (EMNLP-IJCNLP)}, pages 833--844, Hong Kong,
  China. Association for Computational Linguistics.

\bibitem[{Yang et~al.(2019)Yang, Cer, Ahmad, Guo, Law, Constant, Abrego, Yuan,
  Tar, Sung, Strope, and Kurzweil}]{yang2019multilingual}
Yinfei Yang, Daniel Cer, Amin Ahmad, Mandy Guo, Jax Law, Noah Constant,
  Gustavo~Hernandez Abrego, Steve Yuan, Chris Tar, Yun-Hsuan Sung, Brian
  Strope, and Ray Kurzweil. 2019.
\newblock \href {http://arxiv.org/abs/1907.04307} {Multilingual universal
  sentence encoder for semantic retrieval}.

\bibitem[{Zweigenbaum et~al.(2017)Zweigenbaum, Sharoff, and Rapp}]{bucc}
Pierre Zweigenbaum, Serge Sharoff, and Reinhard Rapp. 2017.
\newblock \href {https://doi.org/10.18653/v1/W17-2512} {Overview of the second
  {BUCC} shared task: Spotting parallel sentences in comparable corpora}.
\newblock In \emph{Proceedings of the 10th Workshop on Building and Using
  Comparable Corpora}, pages 60--67, Vancouver, Canada. Association for
  Computational Linguistics.

\end{thebibliography}
\bibliographystyle{acl_natbib}

\appendix

\end{document}